\documentclass[11pt]{article}
\usepackage{coling2020}
\usepackage{times}
\usepackage{url}
\usepackage{latexsym}
\usepackage{amsmath}
\usepackage{graphicx}
\usepackage{adjustbox}
\usepackage{booktabs}
\usepackage{todonotes}
\colingfinalcopy 

\title{Improving Document-Level Sentiment Analysis with \\User and Product Context} 

 \author{Chenyang Lyu \\
   School of Computing \\ 
   Dublin City University \\
   Dublin, Ireland \\
   chenyang.lyu2@mail.dcu.ie \\ \And
   Jennifer Foster \\
   School of Computing \\ 
   Dublin City University \\
   Dublin, Ireland \\
   jennifer.foster@dcu.ie \\ \And
   Yvette Graham \\
   School of Computer Science\\
   \& Statistics \\
   Trinity College Dublin \\
   Dublin, Ireland \\
   ygraham@tcd.ie}

\date{}

\begin{document}
\maketitle
\begin{abstract}
Past work that improves document-level sentiment analysis by encoding user and product information has been limited to considering only the text of 
the current review.
We investigate incorporating additional review text available at the time of sentiment prediction that may prove meaningful for guiding prediction.
Firstly, we incorporate all available historical review text belonging to the author of the review in question.
Secondly, we investigate the inclusion of historical reviews associated with the current product (written by other users).
We achieve this by explicitly storing representations of reviews written by the same user and about the same product
and force the model to \textit{memorize} all reviews for one particular user and product. 
Additionally, we drop the hierarchical architecture used in previous work to enable words in the text to directly attend to each other. Experiment results on IMDB, Yelp 2013 and Yelp 2014 datasets show improvement to state-of-the-art 
of more than 2 percentage points in the best case.
\end{abstract}
\section{Introduction}
%
%
\blfootnote{
    %
    %
    \hspace{-0.65cm}  

    
    \hspace{-0.65cm}  
    This work is licensed under a Creative Commons 
    Attribution 4.0 International Licence.
    Licence details:
    \url{http://creativecommons.org/licenses/by/4.0/}.
    
    %
}



\begin{figure}[htbp]
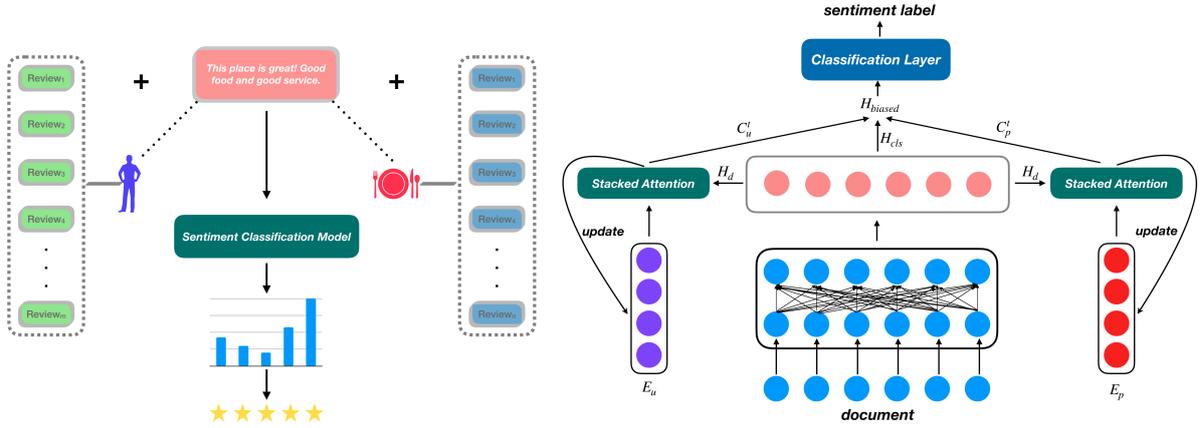


\centering
\includegraphics[scale=0.20]{idea_larger_text.pdf}
\includegraphics[scale=0.20]{model_h_biased.pdf} 

\label{fig-1}
\caption{
Utilizing all historical reviews of corresponding user and products (left); overall architecture of our model, where $E_u$ and $E_p$ are user and product representations (right).}
\end{figure}

Document-level sentiment analysis aims to predict sentiment polarity of text that often takes the form of 
product or service 
reviews.  \newcite{tang-etal-2015-user-product} demonstrated that modelling the individual who has written the review, as well as the product being reviewed, is worthwhile for  polarity prediction, and this has led to exploratory work on how best to combine review text with user/product information in a neural architecture~\cite{chen-etal-2016-neural-sentiment,ma-etal-2017-cascading,dou-2017-capturing,long-etal-2018-dual,amplayo-2019-rethinking,amplayo-etal-2018-cold}. 
A feature common amongst past studies is that user and product IDs are modelled as embedding vectors whose parameters are learned during training. 
We take this idea a step further and represent users and products using the \textit{text of all the reviews belonging to a single user or product} -- see Fig.~\ref{fig-1} (left).

There are two reasons to incorporate review text into user/product modelling. Firstly, the reviews from a given user will reflect their word choices when conveying sentiment. For example, a typical user might use words such as \emph{fantastic} or \emph{excellent} with correspondingly high ratings but another user could use the same words sarcastically with a low rating. Similarly, a group of users 
writing a review of the same product 
may use the same or similar opinionated words to refer to that product.
Secondly,  learning meaningful user and product embeddings that are only updated by back propagation is difficult when a user or product only has a small number of reviews, whereas one may still be able to glean something useful from the text of even a small number of reviews.

A naive approach might compute  representations of all the reviews of a given user or product each time we have a new training sample but this would be too expensive, and we instead propose the following incremental approach: With each new training sample, we obtain the review text representation, with BERT~\cite{bert} as our encoder, before using the representation together with user and product vectors to obtain a user-biased document representation and a product-biased document representation, which are then employed to obtain sentiment polarity. 
We then add the user-biased and product-biased document representations to the corresponding user and product vectors, so that they are ready for the next sample. In doing so, we incrementally store and update  representations of reviews for a given user and product. 
%
%
Unlike \newcite{ma-etal-2017-cascading}, who use a hierarchical structure in which sentence representations are first computed before being combined into a document representation, we let the words in the text directly attend to each other. The architecture we propose 
is depicted on the righthand side of Fig.~\ref{fig-1} and 
is explained in more detail in Section~\ref{sec:3 model}.

We compare performance with a range of systems and results 
show that our approach works, improving on state-of-the-art results for all three benchmark datasets (IMDB, Yelp-13 and Yelp-14).\footnote{http://ir.hit.edu.cn/˜dytang/paper/acl2015/dataset.7z}
We also compare to a version of our own system which does not use the review text representations to encode user and product information. While it performs competitively with other systems, demonstrating the efficacy of our basic architecture, it does not work as well as our proposed system, particularly for reviews written by users or products with only a small number of reviews. 



\section{Methodology}
\label{sec:3 model}
An overview of our model architecture is shown in Figure \ref{fig-1} (right).
The input to our model consists of $d,u,p$, which are the document, the user id and the product id respectively.  $u$ and $p$ are both mapped to embedding vectors, $E_u, E_p$ $\in \mathbf{R^{h}}$. $d$ is fed into the BERT encoder to generate a document representation $H_d \in \mathbf{R^{L\times h}}$ where $L$ is the length of document after tokenization. We then inject $E_u$ and $E_p$, to get the user-product biased document representation $H_{biased}\in \mathbf{R^{h}}$. Finally, we feed the biased document representation $H_{biased}$ into a linear layer followed by a \textit{softmax} layer to get the distribution of the sentiment label $y$. We use \textit{cross-entropy} to calculate the loss between the predictions and ground-truth labels. %
%

\paragraph{Injecting user and product preferences} 
We adopt stacked \textbf{multi-head-attention} $(Q,K,V)$~\cite{transformer} to model the connections between the current document and user/product vectors, which in this work correspond to all historical reviews composed by the user or about the product to date. In a typical dot-product attention $(Q,K,V)$, $Q\in R^{L_Q\times h}$, $K\in R^{L_K\times h}$, $V\in R^{L_V\times h}$. Generally, $L_K = L_v$. $E_u$ and $E_p$ are regarded as queries, $H_d$ as keys and values.
We compute the user-specific document representation, $C_{u}^{t}$, and product-specific document representation, $C_{p}^{t}$ 
as follows:
\vspace{0.1cm}
\begin{align}
\label{stacked_attn}
    C_{u}^{t} = \textbf{stacked-attention}(E_u, H_d, H_d) \quad
    C_{p}^{t} = \textbf{stacked-attention}(E_p, H_d, H_d) 
\end{align}
 where $C_{u}^{t} = attention(C_{u}^{t-1})$, $C_{u}^{0}=E_u$ (similarly for $C_{p}^{t}$), and $t$ is the number of layers of the attention function. In Equation (\ref{stacked_attn}), $C_{u}^{t} \in R^h, C_{p}^{t} \in R^h$.
 
We adopt a \textit{gating mechanism} to obtain importance vectors, $z_u$ and $z_p$, to control the \textit{contribution} of user-specific and product-specific document representations to the output classification:
\vspace{0.1cm}
\begin{align}
    z_u = \sigma(W_{zu}C_{u}^{t} + W_{zh}H_d + b_u) \quad
    z_p = \sigma(W_{zp}C_{p}^{t} + W_{zh}H_d + b_p)
\end{align}
Finally, we obtain the biased document representation $H_{biased}$ by:
\vspace{0.1cm}
\begin{align}
    H_{biased} &= H_{cls} + z_u \odot C_{u}^{t} + z_p \odot C_{p}^{t}
\end{align}
where $H_{cls} \in \mathbf{R^{h}}$ is the final hidden vector of the [CLS] token~\cite{bert} and $ \odot $ is element-wise product. 
\paragraph{Updating the user and product matrix}
To implement our idea of using all reviews composed by $u$ and all reviews about $p$, we incrementally add the current user/product-specific document representation to the corresponding entries in the embedding matrix at each step during training:
\vspace{0.1cm}
\begin{align}
    E_{u}^{'} = \sigma(E_u + \lambda_{u} C_{u}^{t}) \quad    E_{p}^{'} = \sigma(E_p + \lambda_{p} C_{p}^{t})
\end{align}
where $\lambda_u$ and $\lambda_p$ are both learnable real numbers that control the degree to which the representation of the current document should be employed.
\section{Experiments}
\label{sec4: experiments}
\subsection{Experimental Setup}
Our experiments are conducted on the IMDB, Yelp-13 and Yelp-14 benchmark datasets, statistics of which are shown in Table~\ref{data_statistics}.
We use the BERT-base model from HuggingFace ~\cite{Wolf2019HuggingFacesTS}. We train our model with a learning rate chosen from \{8e-6, 3e-5, 5e-5\}, and a weight decay rate chosen from \{0, 1e-1, 1e-2, 1e-3\}, the optimizer we use is AdamW\cite{adamw}. In our experiments, the number of attention layers $t$ is set to 5. The maximum sequence length to BERT is 512. We select the hyper-parameters achieving the best results on the dev set for evaluation on the test set. Evaluation metrics (Accuracy and RMSE) are calculated using scripts from Scikit-learn~ \cite{scikit-learn}.\footnote{https://scikit-learn.org/stable/modules/classes.html\#module-sklearn.metrics}

\begin{table}[!htb]

  \centering
    \begin{tabular}{lrrccccc}
\toprule
    Datasets  & \multicolumn{1}{c}{Classes} & \multicolumn{1}{c}{Documents}  & Users & Products  & Docs/User & Docs/Product  & Words/Doc\\
    \midrule
    IMDB      &      1--10 &  84,919 & 1,310 & 1,635 & 64.82 & 51.94 & 394.6  \\
    Yelp-2013 &       1--5 &  78,966 & 1,631 & 1,633 & 48.42 & 48.36 & 189.3  \\
    Yelp-2014 &       1--5 & 231,163 & 4,818 & 4,194 & 47.97 & 55.11 & 196.9 \\
    \bottomrule
\end{tabular}%
      \caption{Statistics of IMDB, Yelp-2013 and Yelp-2014.}
  \label{data_statistics}%
\end{table}%

\subsection{Results}
 Our experimental results are shown in Table~\ref{exp_res:all}. Our proposed model is named IUPC (\textbf{I}ncorporating \textbf{U}ser-\textbf{P}roduct \textbf{C}ontext). The first two rows 
are baseline models: \textsc{bert vanilla} which is the basic BERT model without user and product information, i.e. only  review text, 
and  
\textsc{IUPC w/o update}, which is the same as our proposed model except that we do not update the user and product embedding matrix by incrementally adding the new review representations. The third row shows our proposed model.
We also compare with results from the 
 NLP-progress leaderboard\footnote{http://nlpprogress.com/english/sentiment\_analysis.html}  of 
 the following models:
 $\\ \\$
\begin{tabular}{cc}
\begin{tabular}{p{6.8cm}}
\textbf{CHIM} \cite{amplayo-2019-rethinking} adopts a chunk-wise matrix representation for user/product attributes; injects user/product information in different locations.
\\

\textbf{CMA} \cite{ma-etal-2017-cascading} A hierarchical LSTM encoding the document; injects user and product information hierarchically. \\

\textbf{DUPMN} \cite{long-etal-2018-dual} encodes the document using a hierarchical LSTM; adopts two memory networks, one for user information and another for product information.
\\
\textbf{HCSC} \cite{amplayo-etal-2018-cold} A combination of CNN and Bi-LSTM as the document encoder; injects user/product information with bias-attention.
\\

\end{tabular} &
\begin{tabular}{p{7.2cm}}

\textbf{HUAPA} \cite{aaai-18-jiajun-chen} adopts two hierarchical models to get user and product specific document representations respectively. 
\\
\textbf{NSC} \cite{chen-etal-2016-neural-sentiment} A hierarchical LSTM encoder incorporating user/ product attributes with word and sentence-level attention. \\

\textbf{RRP-UPM} \cite{cikm19-memory} uses two memory networks besides the user/product embeddings to get refined representations for user/product information.\\

\textbf{UPDMN} \cite{dou-2017-capturing} An LSTM model encoding the document; a memory network capturing user/product information.
\\

\textbf{UPNN} \cite{tang-etal-2015-user-product} adopts a CNN-based encoder and injects user/product information in the embedding and classification layers.
\end{tabular}
\end{tabular}

\begin{table}[!htb]
  \centering
  \begin{tabular}{lcccccccccccccccc}
    \toprule
     && \multicolumn{2}{c}{IMDB} && \multicolumn{2}{c}{Yelp-2013}  && \multicolumn{2}{c}{Yelp-2014} \\ [1ex]
    
     && Acc. (\%) &  RMSE &&  Acc. (\%) &  RMSE &&  Acc. (\%) & RMSE  \\
     
    \midrule

     \textsc{bert vanilla} && 47.9$_\text{0.46}$ & 1.243$_\text{0.019}$ && 67.2$_\text{0.46}$  & 0.647$_\text{0.011}$ && 67.5$_\text{0.71}$  & 0.621$_\text{0.012}$ \\
    
    \textsc{iupc w/o Update} && 52.1$_\text{0.31}$ & 1.194$_\text{0.010}$ && 69.7$_\text{0.37}$  & 0.605$_\text{0.007}$ && 70.0$_\text{0.29}$  & 0.601$_\text{0.007}$ \\
    
    \textsc{iupc} (our model) && 53.8$_\text{0.57}$ & \textbf{1.151$_\text{0.013}$} && \textbf{70.5$_\text{0.29}$}  & \textbf{0.589$_\text{0.004}$} && \textbf{71.2$_\text{0.26}$}  & \textbf{0.592$_\text{0.008}$} \\[1ex]
    \textsc{upnn}  && 43.5 & 1.602 && 59.6  & 0.784 && 60.8  & 0.764  \\
    
    \textsc{updmn} && 46.5 & 1.351 && 63.9  & 0.662 && 61.3  & 0.720  \\
    
    \textsc{nsc} && 53.3 & 1.281 && 65.0  & 0.692 && 66.7  & 0.654  \\
    
    \textsc{cma}   && 54.0 & 1.191 && 66.3  & 0.677 && 67.6  & 0.637  \\
    
    \textsc{dupmn} && 53.9 & 1.279 && 66.2  & 0.667 && 67.6  & 0.639  \\
    
    \textsc{hcsc}  && 54.2 & 1.213 && 65.7  & 0.660 && 67.6  & 0.639  \\
    
    \textsc{huapa} && 55.0 & 1.185 && 68.3  & 0.628 && 68.6  & 0.626  \\
    
    \textsc{chim}  && \textbf{56.4} & 1.161 && 67.8  & 0.641 && 69.2  & 0.622  \\
    
    \textsc{rrp-upm} && 56.2 & 1.174 && 69.0  & 0.629 && 69.1  & 0.621 \\

    \bottomrule
    \end{tabular}%
    \caption{Experimental Results on IMDB, Yelp-2013 and Yelp-2014. Following previous work, we use Accuracy (Acc.) and Root Mean Square Error (RMSE) for evaluation. There are 10 classes in IMDB and 5 classes in Yelp 2013 and Yelp 2014. We run \textsc{bert vanilla}, \textsc{iupc w/o Update} and \textsc{iupc} five times and report the average Accuracy and RMSE. The subscripts represent standard deviation.
    }
  \label{exp_res:all}%
\end{table}
 Our model achieves the best classification accuracy and RMSE on Yelp-2013 and Yelp-2014, and the best RMSE on IMDB. It outperforms previous state-of-the-art results by 1.5 accuracy and 0.042 RMSE on Yelp-2013, by 2.1 accuracy and 0.029 RMSE on Yelp-2014, and by 0.01 RMSE on IMDB. Moreover, it outperforms the two baselines, \textsc{bert vanilla} and \textsc{iupc w/o update} in both classification accuracy and RMSE on all three datasets. 
Although the classification accuracy of our model on IMDB is lower than most of the previous models, we suspect this is because the BERT model is not good at handling longer documents since the input length to BERT is fixed and the average length of documents in IMDB dataset is much longer than the other two datasets.
However, it is worth noting that our model achieves the lowest RMSE which means the predictions of our model are 
\textit{closer} to the gold labels.

\subsection{Analysis}
We analyse the results for reviews whose user or product do not have 
many reviews in the training set and compare our model's performance to the \textsc{iupc w/o update} baseline for one dataset (Yelp-2013 dev). 
We select only reviews where the number of reviews by that user or for that product falls below three thresholds: {40\%, 60\%, 80\%}, where \% stands for the number of reviews for a given user/product relative to the average number of reviews for all users/products. Table~\ref{exp_res:filtered} shows that our model performs better than \textsc{iupc w/o update} 
when there are only a small number of previous reviews available for a given product/user. In other words, when a user or product does not have many reviews, its \textsc{iupc w/o update} embedding which is only updated by gradient descent, cannot capture user/product preference as well as our model which explicitly takes advantage of historical review text in its user/product representations.

\begin{table}[!htb]

  \centering
    \begin{tabular}{lcccccccccc}
    \toprule
     && \multicolumn{2}{c}{40\%} && \multicolumn{2}{c}{60\%}  && \multicolumn{2}{c}{80\%} \\ 
    
     && Acc. (\%) & RMSE && Acc. (\%) & RMSE  && Acc. (\%) & RMSE   \\
    \midrule
    
    \textsc{iupc w/o Update} && 63.0 & 0.608 && 64.0  & 0.665 && 66.8  & 0.643 \\
    
    \textsc{iupc} (our model) && \textbf{65.7} & \textbf{0.585} && \textbf{66.8}  & \textbf{0.649} && \textbf{67.9}  & \textbf{0.631}  \\
    \bottomrule
    \end{tabular}%
      \caption{Analysis of three lower-resource scenarios where \% denotes a threshold filter corresponding to the proportion of reviews available relative to the average number in the dataset Yelp-2013 (dev).}
  \label{exp_res:filtered}%
\end{table}%

In order to get a better idea of where there is room for improvement for \textsc{iupc}, we examine the 43  Yelp-13 dev set cases, where the predicted label differs from the gold label by more than two points. There are a handful of cases of sarcasm, e.g. \textit{that \textbf{lovely} tempe waste/tap water taste in the food}, but the most noteworthy phenomenon is mixed sentiment, e.g. \textit{tacos were good the soup was not tasty}, or the more subtle \textit{brave the scary parking and lack of ambiance}. It is not always clear from the reviews which aspect of the service the rating is directed towards. This suggests that  aspect-based sentiment analysis~\cite{pontiki-etal-2014-semeval} might be useful here, and training an \textsc{iupc} model for this task is a possible avenue for future work.

\section{Conclusion}
In this paper, we propose a neural sentiment analysis architecture that explicitly utilizes all past reviews from a given user or product to improve sentiment polarity classification on the document level. Our experimental results on the IMDB, Yelp-13 and Yelp-14 datasets demonstrate that incorporating this additional context
is effective,
particularly for the Yelp datasets. The code used to run the experiments is available for use by the research community.\footnote{https://github.com/lyuchenyang/Document-level-Sentiment-Analysis-with-User-and-Product-Context}


\section*{Acknowledgements}
This work was funded by Science Foundation Ireland through the SFI Centre for Research Training in Machine Learning (18/CRT/6183). We also thank the  reviewers for their helpful comments.

\bibliographystyle{coling}
\bibliography{coling2020}

\end{document}